\documentclass{article}
\usepackage[sglblindworkshop,final]{neurips_2026}
\workshoptitle{Vision-Language Models for Real-World Deployment (VLM4RWD), VistaQA Challenge Track 1}

\usepackage[utf8]{inputenc}
\usepackage[T1]{fontenc}
\usepackage[hidelinks]{hyperref}
\usepackage{url}
\usepackage{booktabs}
\usepackage{array}
\usepackage{amsmath}
\usepackage{amssymb}
\usepackage{microtype}
\usepackage{xspace}
\usepackage{pifont}
\usepackage{inconsolata}
\usepackage{xcolor}
\usepackage{colortbl}
\usepackage{fontawesome5}
\makeatletter\providecommand\insert@pcolumn{\insert@column}\makeatother
\usepackage{tikz}
\usepackage{graphicx}
\usetikzlibrary{positioning,arrows.meta,fit,shadows.blur}
\usepackage[skins,breakable]{tcolorbox}

\definecolor{gain}{HTML}{1E8F7A}
\definecolor{loss}{HTML}{C2417A}
\definecolor{muted}{HTML}{5B6474}
\definecolor{lavender}{HTML}{A7A8F0}
\definecolor{lavtint}{HTML}{ECECFC}
\definecolor{boxtitle}{HTML}{8180DE}
\definecolor{boxback}{HTML}{F5F5FE}
\definecolor{repolink}{HTML}{367DBD}
\newcommand{\ours}{\rowcolor{lavtint}}

\newcommand{\vs}{\textit{vs.}\@\xspace}

\DeclareRobustCommand{\micon}[1]{\raisebox{-0.15\height}{\includegraphics[height=1.55ex]{logos/#1.pdf}}\hspace{0.28em}}
\newcommand{\up}[1]{\ensuremath{_{\textcolor{gain}{+#1}}}}
\newcommand{\dn}[1]{\ensuremath{_{\textcolor{loss}{-#1}}}}
\newcommand{\gup}[1]{\textcolor{gain}{$+$#1}}
\newcommand{\gdn}[1]{\textcolor{loss}{$-$#1}}

\newcommand{\pct}[1]{\hspace{0.5em}\makebox[2.1em][r]{\textcolor{muted}{#1\%}}}
\newcommand{\agree}{\textcolor{gain}{\ding{51}}}
\newcommand{\disagree}{\textcolor{loss}{\ding{55}}}
\newtcolorbox{promptbox}[1]{enhanced, breakable, colback=boxback, colframe=lavender, boxrule=0.6pt, arc=2pt,
  left=6pt, right=6pt, top=4pt, bottom=4pt, fonttitle=\bfseries\small, coltitle=white, colbacktitle=boxtitle,
  before title={\hypersetup{citecolor=white,linkcolor=white}}, title={#1}, title after break={#1 (continued)},
  fontupper=\small, fontlower=\small, before upper={\setlength{\parindent}{0pt}\setlength{\parskip}{3pt}},
  before lower={\setlength{\parindent}{0pt}\setlength{\parskip}{3pt}},
  segmentation style={draw=lavender, line width=0.5pt, dashed}, before skip=4pt, after skip=6pt}
\newcommand{\ph}[1]{\textit{$\langle$#1$\rangle$}}
\newcommand{\pipelineversion}{3}

\colorlet{ishraq}{black}
\newcommand{\ishraq}[1]{{\color{ishraq}#1}}

\title{Grounding the Ungroundable:\\ Answers, Masks, and Absences in Zero-Shot VQA}

\author{%
  {\mdseries\large Md Farhan Ishraq$^{1}$, Didhiti Nahid$^{1}$, Md Farhan Ishmam$^{1,2}$, Syed Rifat Raiyan$^{1}$} \\[4pt]
  {\mdseries\large $^{1}$Islamic University of Technology \quad $^{2}$University of Utah} \\[4pt]
  \ishraq{{\small\faGithub\;\href{https://github.com/farhanishraq17/vistaqa-zero-shot-grounding}{\textcolor{repolink}{\texttt{farhanishraq17/vistaqa-zero-shot-grounding}}}}}
}

\begin{document}
\maketitle

\begin{abstract}
A hallmark of a deployable vision-language system is that it answers a question about an image, grounds the answer in
pixels, and recognizes when the object asked about is absent. The last of these remains fragile in the strongest published
systems, which pair one vision-language model with a promptable segmenter and trust that model alone to notice a false
premise. We present a zero-shot chain of open-weight models that answers, grounds, and then doubts the premise, and evaluate
it on VistaQA, where 27\% of questions ask about something absent. Qwen3.6-27B answers and boxes its evidence, SAM3 turns the
boxes into masks, a separate Qwen3-VL-8B pass, blind to the answer, checks the premise, and a guarded rule raises
under-counted answers. We made all configuration decisions using an external 560-question proxy set and finalized the setup before evaluating VistaQA. With the AWQ build of the benchmark's judge, our entry scores \ishraq{54.65} GROVE
(\ishraq{54.43} with the unquantized model) against 45.53 for the best published system (GPT-5.4-Thinking + SAM3), mostly
through false-premise questions (\ishraq{75.2}~\vs~47.9). The lesson for deployment is that a grounded answering system gains
most of its edge by recognizing when there is nothing to ground.
\end{abstract}

\vspace{-4pt}%
\section{Introduction}
\vspace{-2pt}%
\ishraq{Grounding an answer presumes that there is something to ground. VistaQA~\cite{vistaqa} puts that presumption to the
test: it asks a vision-language model (VLM) to answer a question about an image and to segment every instance of
its evidence, yet 27\% of its 1,157 questions rest on a false premise. For these, correctness requires not only identifying what is present, but also recognizing what is absent. Its metric, GROVE
(Section~\ref{sec:setup}), caps a wrong answer at $0.32$ or less and gives such a response the full 1.0, so answer accuracy and a
reliable premise check matter more than mask quality. Yet the strongest published systems pair one VLM with
SAM3~\cite{sam3} and leave the premise to that model alone, assuming that a model intent on answering will also notice.}

\ishraq{We introduce a training-free chain of open-weight models that separates three jobs (Figure~\ref{fig:pipe},
Section~\ref{sec:system}): one VLM answers and boxes its evidence, a promptable segmenter turns the boxes into masks, and a
second VLM pass, blind to the answer, checks the premise. Our contributions are:
\begin{list}{\textbullet}{\setlength{\leftmargin}{1.4em}\setlength{\labelwidth}{0.8em}\setlength{\labelsep}{0.5em}%
\setlength{\itemsep}{0pt}\setlength{\parsep}{0pt}\setlength{\topsep}{1pt}\setlength{\partopsep}{0pt}}
\item \textbf{A zero-shot grounded answering system.} We introduce a zero-shot pipeline that decouples answer generation, pixel-level evidence extraction, and false-premise verification across open-weight models scoring 54.65 GROVE on VistaQA
against 45.53 for the best published system.
\item \textbf{A benchmark-isolated zero-shot protocol.} We develop and select our system using external proxy datasets, keeping VistaQA untouched throughout development to ensure a more reliable assessment of zero-shot generalization.
\item \textbf{An empirical analysis of generalization.} We show that false-premise verification consistently transfers to VistaQA, while several backbone and reasoning enhancements do not.
\end{list}}

\vspace{-4pt}%
\section{Related work}
\vspace{-2pt}%
\textbf{Grounded answering with masks.} Reasoning-segmentation models such as LISA~\cite{lisa}, GLaMM~\cite{glamm} and
Sa2VA~\cite{sa2va} train a VLM to emit segmentation tokens for a mask decoder. On VistaQA, the nine published systems
without SAM3, these three included, score 11.8 to 29.0 GROVE, whereas the five that pair a general VLM with SAM3 score 39.6
to 45.5 (Table~\ref{tab:main}). Like Grounded-SAM~\cite{groundedsam}, our system belongs to this modular family: a VLM or
detector proposes boxes, and a promptable segmenter (SAM~\cite{sam}, SAM~2~\cite{sam2} or SAM3) draws the masks.

\vspace{-1.5pt}%
\textbf{False premises and hallucination.} POPE~\cite{pope} measures object hallucination with yes/no presence questions;
we reuse its form to confirm that an object is missing. SESAME~\cite{sesame} trains a multimodal model to recognize and
correct false premises before segmenting; we obtain similar behavior without training, from a check that asks only what the
question assumes and the image lacks.

\vspace{-4pt}%
\section{Methodology}\label{sec:system}
\vspace{-2pt}%
\begin{figure}[t]
\centering
\ifcase\pipelineversion\or
\input{fig_pipeline}
\vspace{-3pt}%
\caption{Passes A--D run independently, and fixed rules merge their outputs. The example is illustrative (a synthetic
image and a question not from VistaQA); C finds nothing missing, and D does not apply.}
\or
\includegraphics[width=\textwidth]{diagrams/pipeline_alt.pdf}
\vspace{-3pt}%
\caption{The entry's four passes and three merge rules. A answers in one sentence and boxes its evidence, B turns each box
into a SAM3 mask, C names what the question $q$ assumes but the image $x$ lacks, without seeing the answer, and confirms it
with a yes/no question, and D can only raise the count on plain counting questions. For $q_1$, B's mask joins A's answer
(rule~1); for $q_2$, the confirmed absence replaces A's answer (struck through) and drops its masks (rule~2). The scene is
synthetic, and both questions are illustrative rather than drawn from VistaQA.}
\or
\includegraphics[width=\textwidth]{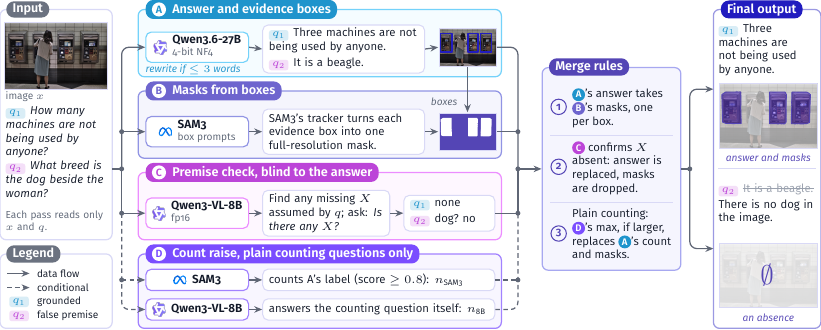}
\vspace{-3pt}%
\caption{\ishraq{\textbf{\textit{Overview of the proposed zero-shot pipeline:}} (A) generates answers and evidence boxes, (B) converts boxes into SAM3 masks, (C) independently verifies false premises, and (D) estimates counts. Fixed merge rules yield grounded answers with masks or absence predictions. The image and $q_1$ are from VistaQA sample 811 (CC BY 4.0); $q_2$ and its outputs are illustrative.}}
\fi
\label{fig:pipe}
\vspace{-6pt}%
\end{figure}

\vspace{-1.5pt}%
\ishraq{\textbf{Problem formulation.} Given an image $x\in\mathcal{I}$ of size $H\times W$ and a question
$q\in\mathcal{Q}$, a system returns $\langle a,\mathcal{M}\rangle$: a free-form answer $a$ and a set
$\mathcal{M}=\{m_i\}_{i=1}^{n}$ of instance masks $m_i\in\{0,1\}^{H\times W}$ of its evidence. Under a false premise, $a$
must state the absence and $\mathcal{M}=\emptyset$.}

\vspace{-1.5pt}%
\ishraq{\textbf{A. Answer and evidence boxes.} The answerer $f_A:(x,q)\mapsto\langle a_A,\mathcal{B}\rangle$,
Qwen3.6-27B~\cite{qwen3} in 4-bit NF4, is prompted once (Appendix~\ref{app:prompts}) for a one-sentence answer $a_A$ and a
set $\mathcal{B}=\{(\ell_j,b_j)\}$ of labeled boxes, one per evidence instance, with $\mathcal{B}=\emptyset$ if the object is
absent or $q$ unanswerable. As the judge rejects bare answers, $f_A$ rewrites any $a_A$ of at most three words
into a sentence.}

\vspace{-1.5pt}%
\ishraq{\textbf{B. Masks.} SAM3's tracker head $g$ maps each box to a full-size mask,
$\mathcal{M}_A=\{g(x,b_j)\}_j$. On these boxes, SAM2.1-L tied on the proxy (71.74 \vs 71.72); we kept SAM3 for the count
raise (D).}

\vspace{-1.5pt}%
\ishraq{\textbf{C. Premise check.} The checker $f_C:(x,q)\mapsto X$, Qwen3-VL-8B~\cite{qwen3vl}, never sees $a_A$: told
not to answer, it names any object, person, text or detail $X$ that $q$ assumes and $x$ lacks, or $X=\emptyset$. A
POPE-style yes/no question $v(x,X)$ confirms it; on ``no'', the output becomes
$\langle\text{``There is no $X$ in the image.''},\emptyset\rangle$. We chose the 8B because, at similar recall, the 27B
raised more proxy false alarms (36 \vs 17).}

\vspace{-1.5pt}%
\ishraq{\textbf{D. Count raise.} All our counters under-count. On plain counting questions (at most 12 words, no
arithmetic), let $n_A$ be the count in $a_A$, $n_S$ SAM3's count of A's evidence label at score
$\geq\tau=0.8$, and $n_{8\mathrm{B}}$ the 8B's count. If $n^\star=\max(n_S,n_{8\mathrm{B}})>n_A$, $a$ takes $n^\star$ and
$\mathcal{M}$ the winner's masks. The guard comes from a MathVista~\cite{mathvista} count probe, where unguarded raises
broke more than they fixed.}

\newpage
\vspace{-4pt}%
\section{Experimental setup}\label{sec:setup}
\vspace{-2pt}%
\ishraq{\textbf{Benchmark and metric.} VistaQA's metric, GROVE, scores a system $\pi$ on a set of $N$ samples as
\mbox{$G(\pi)=\tfrac{100}{N}\sum_{i}\sqrt{\max(T_i,\epsilon)\,\max(S_i,\epsilon)}$}, where $T_i\in\{0,1\}$ is
an LLM judge's verdict on answer $i$ and $S_i$ the sum of Hungarian-matched IoUs between the predicted masks $\mathcal{M}_i$
and the ground-truth masks $\mathcal{M}_i^{\star}$, divided by $\max(|\mathcal{M}_i|,|\mathcal{M}_i^{\star}|)$.
Table~\ref{tab:main} compares systems at the benchmark's floor $\epsilon=0.1$; scores not marked so use the code's
default, $\epsilon=0.01$. The
benchmark's judge is Qwen2.5-14B-Instruct~\cite{qwen25}; the entry's score uses a compressed copy of it (its AWQ build),
and we also report the uncompressed (full) judge, having committed to both before running either. Each difference
$\Delta=G(\pi)-G(\pi')$ has a 95\% confidence interval (CI) from random redraws of the samples.}

\vspace{-1.5pt}%
\ishraq{\textbf{Research questions.} Let $\pi_k$ denote the system after pass $k$, so that $\pi_D$ is the entry, and
$\mathcal{P}$ the 14 published systems. \textbf{RQ1}: does the entry lead the published board,
$G(\pi_D)>\max_{p\in\mathcal{P}}G(p)$? \textbf{RQ2}: what does each later stage add, $\Delta_C=G(\pi_C)-G(\pi_B)$ and
$\Delta_D=G(\pi_D)-G(\pi_C)$? \textbf{RQ3}: does a gain on the proxy predict one on VistaQA,
$\operatorname{sign}\Delta^{\mathrm{proxy}}=\operatorname{sign}\Delta^{\mathrm{VistaQA}}$?}

\vspace{-1.5pt}%
\ishraq{\textbf{Zero-shot protocol and development data.} The final entry was evaluated on VistaQA only after all
development decisions had been finalized; all design choices were made on an external proxy set comprising 560 templated
questions with masks from COCO val2017~\cite{coco}, PACO-LVIS~\cite{paco}, and COCO-Text v2~\cite{cocotext}, including 150
(26.8\%) false-premise questions, alongside text-only probes from MathVista and TextVQA~\cite{textvqa}
(Appendix~\ref{app:dev}). The frozen configuration (73.81 proxy score) and its predictions were registered in version
control before VistaQA evaluation and judging, respectively. Each pass accesses only the image and question, and the
evaluation harness was validated before scoring (Appendix~\ref{app:judge}). The only post-freeze modification was a
formatting fix: three masks were resized to the metadata size of two VistaQA images whose files are smaller than
declared.}

\vspace{-1.5pt}%
\ishraq{\textbf{Implementation.} Greedy decoding yielded 147 identical answers across 150 VistaQA samples, while SAM3 reproduced 259 masks with a mean IoU of 0.9998.}

\vspace{-1.5pt}%

\begin{table}[t]
\color{ishraq}
\centering\scriptsize
\setlength{\tabcolsep}{1.4pt}
\caption{GROVE on VistaQA: 14 published systems (quoted from the VistaQA paper, its appendix for $\epsilon=0.01$) and our
entry under the AWQ judge (registered) and the unquantized judge (full). Gr.: grounded; FP: false premise; bold: best per
column; $\Delta$: entry minus best published.}
\label{tab:main}
\vspace{-2pt}%
\begin{tabular}{l*{16}{r}}
\toprule
 & \multicolumn{1}{c}{$\epsilon{=}.01$} & \multicolumn{15}{c}{$\epsilon=0.1$ (the benchmark paper's setting)} \\
\cmidrule(lr){2-2}\cmidrule(l){3-17}
\textbf{System} & \textbf{All} & \textbf{All} & \textbf{Ident.} & \textbf{Attr.} & \textbf{OCR} & \textbf{Spat.} & \textbf{Count} & \textbf{Comp.}
 & \textbf{Gr.} & \textbf{FP} & \textbf{Indoor} & \textbf{Outd.} & \textbf{Math} & \textbf{AV} & \textbf{Robot.} & \textbf{Sci.} \\
\midrule
\multicolumn{17}{l}{\textit{General VLM + SAM3}} \\
\micon{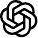}GPT-5.4-Thinking & 35.86 & 45.53 & 53.62 & 44.73 & 44.16 & 38.86 & \textbf{52.21} & 38.93 & 44.65 & 47.89 & 45.39 & 45.92 & \textbf{41.51} & 53.18 & \textbf{47.01} & 39.24 \\
\micon{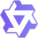}Qwen3-VL-4B & 34.14 & 43.30 & 50.36 & 47.46 & 40.96 & 36.32 & 45.52 & 38.43 & 36.81 & 60.72 & 43.64 & 46.84 & 27.37 & 59.39 & 41.78 & 36.80 \\
\micon{qwen}Qwen3-VL-32B & 33.51 & 42.72 & 48.41 & 46.37 & 42.14 & 35.91 & 45.72 & 37.41 & 37.63 & 56.41 & 44.27 & 47.63 & 23.23 & 58.40 & 40.09 & 37.85 \\
\micon{gpt}GPT-5.4 & 31.61 & 41.50 & 48.22 & 40.43 & 41.33 & 35.62 & 47.06 & 35.86 & 43.28 & 36.72 & 42.06 & 46.05 & 28.99 & 53.43 & 39.87 & 35.54 \\
\micon{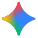}Gemini 3 & 25.49 & 39.63 & 43.10 & 40.11 & 42.55 & 33.53 & 42.81 & 35.84 & 33.48 & 56.14 & 39.37 & 39.40 & 31.40 & 48.42 & 37.15 & 39.87 \\
\midrule
\multicolumn{17}{l}{\textit{Other published systems (no SAM3)}} \\
VRT-RL & 17.93 & 29.02 & 34.92 & 28.74 & 27.43 & 24.44 & 29.94 & 28.01 & 32.09 & 20.79 & 29.61 & 33.06 & 24.68 & 30.76 & 27.74 & 27.27 \\
Sa2VA-8B & 16.02 & 25.74 & 26.79 & 30.44 & 25.36 & 22.87 & 20.46 & 28.36 & 31.01 & 11.58 & 22.64 & 24.03 & 14.87 & 36.87 & 25.10 & 28.09 \\
UniPixel-7B & 12.02 & 22.39 & 22.24 & 28.29 & 22.19 & 19.90 & 15.36 & 26.34 & 25.98 & 12.75 & 21.76 & 23.20 & 12.90 & 28.87 & 21.02 & 24.15 \\
TreeVGR-7B & 7.54 & 19.03 & 18.28 & 18.55 & 16.42 & 17.90 & 20.25 & 22.65 & 17.68 & 22.66 & 18.41 & 20.45 & 12.70 & 23.51 & 19.76 & 17.78 \\
SESAME-7B & 5.87 & 17.02 & 17.90 & 17.14 & 13.62 & 18.48 & 17.60 & 16.90 & 12.17 & 30.04 & 17.20 & 17.68 & 11.05 & 20.77 & 15.92 & 17.97 \\
LISA-7B & 3.61 & 15.47 & 15.16 & 16.92 & 12.44 & 15.29 & 16.56 & 16.23 & 14.01 & 19.39 & 16.01 & 16.41 & 12.15 & 16.50 & 15.97 & 14.95 \\
GLaMM-7B & 2.93 & 13.53 & 14.86 & 12.89 & 12.34 & 14.19 & 11.84 & 14.70 & 14.48 & 10.96 & 12.82 & 13.75 & 11.13 & 15.70 & 13.43 & 13.72 \\
LaSagnA-7B & 2.37 & 12.35 & 13.31 & 12.03 & 11.04 & 13.23 & 11.14 & 13.05 & 13.12 & 10.28 & 11.31 & 12.18 & 11.07 & 14.05 & 12.09 & 13.07 \\
Uground-7B & 2.13 & 11.78 & 10.95 & 11.69 & 11.15 & 11.57 & 12.15 & 13.22 & 10.94 & 14.03 & 11.32 & 10.74 & 10.78 & 11.79 & 13.40 & 12.38 \\
\midrule
\ours Ours$_\text{AWQ}$ (entry) & \textbf{48.40} & \textbf{54.65} & 56.75 & 58.59 & \textbf{63.79} & \textbf{48.28} & 52.15 & 49.21 & 46.98 & \textbf{75.24} & \textbf{59.71} & 60.77 & 34.76 & 71.87 & 40.08 & \textbf{55.57} \\
\ours Ours$_\text{full}$ & 48.14 & 54.43 & \textbf{57.32} & \textbf{59.06} & 63.04 & 47.15 & 50.21 & \textbf{50.50} & \textbf{47.86} & 72.05 & 59.04 & \textbf{61.06} & 35.85 & \textbf{72.53} & 38.14 & 55.12 \\
\addlinespace[1pt]
$\Delta$ to best published & \gup{12.5} & \gup{9.1} & \gup{3.1} & \gup{11.1} & \gup{19.6} & \gup{9.4} & \gdn{0.1} & \gup{10.3} & \gup{2.3} & \gup{14.5} & \gup{14.3} & \gup{13.1} & \gdn{6.8} & \gup{12.5} & \gdn{6.9} & \gup{15.7} \\
\bottomrule
\end{tabular}
\vspace{-6pt}%
\end{table}

\vspace{-4pt}%
\section{Experiments and results}\label{sec:results}
\vspace{-2pt}%
\ishraq{\textbf{RQ1: the entry leads the published board.} It is first under both judges (Table~\ref{tab:main}): 54.65
(95\% interval 52.5--56.8) against 45.53, and 48.40 against 35.86~\cite{vistaqa} at the evaluation code's $\epsilon=0.01$.
It leads 11 of the 14 category columns, is level on counting, and trails GPT-5.4-Thinking on math and robotics. Against
that system, text accuracy is level (60.76 \vs 61.02): higher on false-premise questions (72.0\% \vs 39.2\%) and lower on
grounded ones (56.6\% \vs 69.2\%). Our masks are better (mIoU 52.05 \vs 41.98). Published per-sample outputs are
unreleased, so no comparison is paired.}

\begin{table}[t]
\centering\small
\setlength{\tabcolsep}{4.5pt}%
\caption{\textbf{Stage-wise ablation results on the proxy set and VistaQA.} The proxy guided all development decisions; the first two stages were evaluated on VistaQA for analysis only. Bold: best per column.}
\label{tab:abl}
\vspace{-2pt}%
\begin{tabular}{llllll}
\toprule
\textbf{Stage} & \ishraq{\textbf{Proxy}} & \ishraq{\textbf{VistaQA}$_{\epsilon=0.1}$} & \textbf{VistaQA}$_\text{AWQ}$ & \textbf{VistaQA}$_\text{full}$ & \textbf{Text acc.}$_\text{AWQ}$ \\
\midrule
A + B: answer, boxes, masks & 65.07 & \ishraq{51.78} & 45.46 & 45.85 & 59.55 \\
+ C: premise check & 72.27\up{7.20} & \ishraq{\textbf{55.11}}\up{3.33} & \textbf{48.95}\up{3.49} & \textbf{48.62}\up{2.77} & \textbf{61.45}\up{1.90} \\
\ours + D: count raise (entry) & \textbf{73.81}\up{1.54} & \ishraq{54.65}\dn{0.46} & 48.40\dn{0.55} & 48.14\dn{0.48} & 60.76\dn{0.69} \\
\bottomrule
\end{tabular}
\vspace{-6pt}%
\end{table}

\vspace{-1.5pt}%
\ishraq{\textbf{RQ2: the premise check adds $\Delta_C=+3.49$, and the count raise loses $\Delta_D=-0.55$
(Table~\ref{tab:abl}).} The check (95\% CI $+1.65$ to $+5.35$) caught 222 of 314 false-premise questions but also fired on
110 of 843 grounded ones (13\%; 4\% on the proxy). The raise (CI $-1.02$ to $-0.13$) added 1.5 on the proxy; on VistaQA, 10
of 23 raised answers turned wrong and 2 turned right.}
A bug found after scoring, an 8B count pass run without masks, cost 0.07 of this loss (Appendix~\ref{app:anat}).

\vspace{-1.5pt}%
\ishraq{\textbf{RQ3: a proxy gain only partly predicts a VistaQA gain.} The dev sets gave the right sign for 5 of 12
changes (Appendix~\ref{app:transfer}). The premise check and the 8B weight edit
(Appendix~\ref{app:wsp}) were the two improvements that held, both focused on false premises.}

\vspace{-1.5pt}%
\textbf{Sources of score loss.} Of the 51.6 points lost \ishraq{at $\epsilon=0.01$}, 30.4 come from wrong grounded answers,
11.9 from imperfect masks on right ones, 7.3 from wrong false-premise answers and 1.9 from missing or extra masks. Of 460
right grounded answers with masks, 169 have more masks than the ground truth (\ishraq{the proxy}: 1 of 310);
Appendix~\ref{app:mv} tests a Set-of-Mark fix.

\vspace{-4pt}%
\section{Discussion and limitations}\label{sec:disc}
\vspace{-2pt}%
Our system splits one judgment into two: what the answer is, and whether the image holds what the question takes for
granted. Published systems leave both to one model and tacitly trust it to notice a false premise. While
answering, Qwen3-VL-8B rejects only 68 of the proxy's 150 false premises; asked only what the question assumes, the same
model catches 138 (Appendix~\ref{app:wsp}). Run as a separate check, blind to the answer, the 8B added 7.20 points on the
proxy and 3.49 on VistaQA, the one stage of the entry whose gain held on both.

\vspace{-1.5pt}%
A proxy can be a microcosm of VistaQA's tasks and false-premise share (26.8\% \vs 27.1\%) and still differ in its images.
Both gains that held target false premises (RQ3), and we surmise that absence travels better than appearance: a missing
object poses much the same problem on COCO as on VistaQA, whereas a new backbone or prompt meets images the proxies only
approximate. Four later systems (v2--v5), each chosen on dev sets and registered before its single VistaQA run, landed
within 0.4 points of the entry, although v2 gained 8.4 points on a second, more varied proxy (Appendix~\ref{app:v2}). A
large dev-set gain did not portend a VistaQA gain, and the dev sets also misjudged both model swaps, one in each
direction (Appendix~\ref{app:ms}).

\vspace{-1.5pt}%
The entry still trails GPT-5.4-Thinking on math and robotics, by 6.8 and 6.9 points at $\epsilon=0.1$
(Table~\ref{tab:main}), and the premise check costs 4.6 points on grounded questions. The proxy understated what VistaQA
exposed (Appendices~\ref{app:qual} and~\ref{app:anat}): the check's false alarms (4\% \vs 13\% of grounded questions),
over-masking, and the count raise, which gained 1.54 on the proxy's templated counts but lost 0.55 on VistaQA. We scored 22
systems on VistaQA without multiple-comparison correction; only the entry was frozen before VistaQA was first scored.

\vspace{-4pt}%
\section{Conclusion}
\vspace{-2pt}%
\ishraq{We introduce a training-free chain of open-weight models that answers a question, grounds the answer in masks, and
separately checks the question's premise. Developed entirely on external data, it scores 54.65 GROVE on VistaQA against
45.53 for the best of 14 published systems, with most of the margin on false-premise questions.} The only dev-set gains
that held on VistaQA came from the premise check and a weight edit of the 8B, both aimed at false premises; a newer
backbone, thinking and a calculator brought none. \ishraq{We provide the code, run log, predictions, and official
evaluation at our GitHub repository \url{https://github.com/farhanishraq17/vistaqa-zero-shot-grounding}. A system built to
ground its answers must also recognize when there is nothing to ground.}

\bibliographystyle{plain}
\bibliography{refs}

\appendix

\section{\texorpdfstring{\ishraq{Prompt templates}}{Prompt templates}}\label{app:prompts}
\ishraq{The following boxes contain the prompts used in the released implementation, with \ph{\dots} indicating runtime-filled slots. The first three boxes belong to the entry, and the remaining four serve the analysis systems of
Appendices~\ref{app:v2} and~\ref{app:mv}.}

\begin{promptbox}{Pass A: answer and evidence boxes (Qwen3.6-27B)}
You are given an image and a question about it.\\
1. Answer the question accurately and concisely, as one short complete sentence.\\
2. Localize the visual evidence: every object or region in the image that your answer is based on, one entry per object
instance, each with a bounding box [x1, y1, x2, y2] in coordinates normalized to 0-1000.\\
- If you count objects, give one box per counted object; the number of boxes must equal your count.\\
- For questions about text, box the text region you read.\\
- If the object or entity the question asks about does not exist in the image, or the question cannot be answered from the
image, say so in the answer and return an empty evidence list.

Question: \ph{question}

Return only JSON: \texttt{\{"answer": "...", "evidence": [\{"label": "...", "bbox\_2d": [x1, y1, x2, y2]\}]\}}
\end{promptbox}

\begin{promptbox}{Short-answer rewrite, for answers of at most three words (Qwen3.6-27B)}
Question: \ph{question}\\
Short answer: \ph{answer}\\
Rewrite the short answer as one complete sentence that restates what the question asks about. Do not add or change any
information. Return only the sentence.
\end{promptbox}

\begin{promptbox}{Pass C: premise check, then its yes/no confirmation when an item is named (Qwen3-VL-8B)}
Question about this image: \ph{question}\\
Do not answer the question. Check its premise: does the question assume an object, person, text or detail that is not
actually visible in the image? Look carefully at the whole image.\\
If something it assumes is missing, name it in a few words; otherwise answer none.\\
Return only JSON: \texttt{\{"missing": "..."\}}
\tcblower
Is there any \ph{missing} in the image? Answer yes or no.
\end{promptbox}

\begin{promptbox}{Appendix~\ref{app:v2}: figure router (Qwen3-VL-8B), then reasoning router (Qwen3.8-27B, no thinking)}
Is this image a chart, plot, table, diagram or geometric figure, rather than a photograph or a rendering of a real scene?
Answer yes or no.
\tcblower
Question about this image: \ph{question}\\
Do not answer it. Does answering it need several steps of calculation or reasoning (arithmetic, algebra, geometry, physics, a
logic puzzle), rather than reading a value, label or count directly from the image? Answer yes or no.
\end{promptbox}

\begin{promptbox}{Appendix~\ref{app:v2}: evidence prompt for routed figures (Pass A with these lines in place of its step 2)}
2. Localize the visual evidence: only what directly shows the answer, one entry per object instance, each with a bounding box
[x1, y1, x2, y2] in coordinates normalized to 0-1000.\\
- Box the answer element itself, not what you read to find it: the bar or slice rather than its label or the legend, the
table cell holding the value rather than its row and column headers, the shape rather than its side lengths.\\
- If the question compares or relates objects, box each of them.
\end{promptbox}

\begin{promptbox}{Appendix~\ref{app:v2}: calculator of v4 (Qwen3.8-27B, no thinking)}
Question about this image: \ph{question}\\
Write a short Python program that computes the answer. First read the values you need from the image and assign them to
variables, with a comment saying where each comes from; then compute; then print the final answer exactly as it should be
given: a number (with its unit if the question asks for one) or the text of the correct option. Use only arithmetic and the
math module. Reply with the program in a \texttt{\textasciigrave\textasciigrave\textasciigrave python} block.
\end{promptbox}

\begin{promptbox}{Appendix~\ref{app:mv}: Set-of-Mark mask check (Qwen3.6-27B; Qwen3-VL-8B compared)}
Question about this image: \ph{question}\\
Answer: \ph{answer}\\
The image shows \ph{n} numbered boxes. Which numbered boxes mark visual evidence required to support this answer? Reply with a
JSON list of box numbers only.
\end{promptbox}

\section{Qualitative examples}\label{app:qual}
\begin{figure}[!htbp]
\centering
\includegraphics[width=\textwidth]{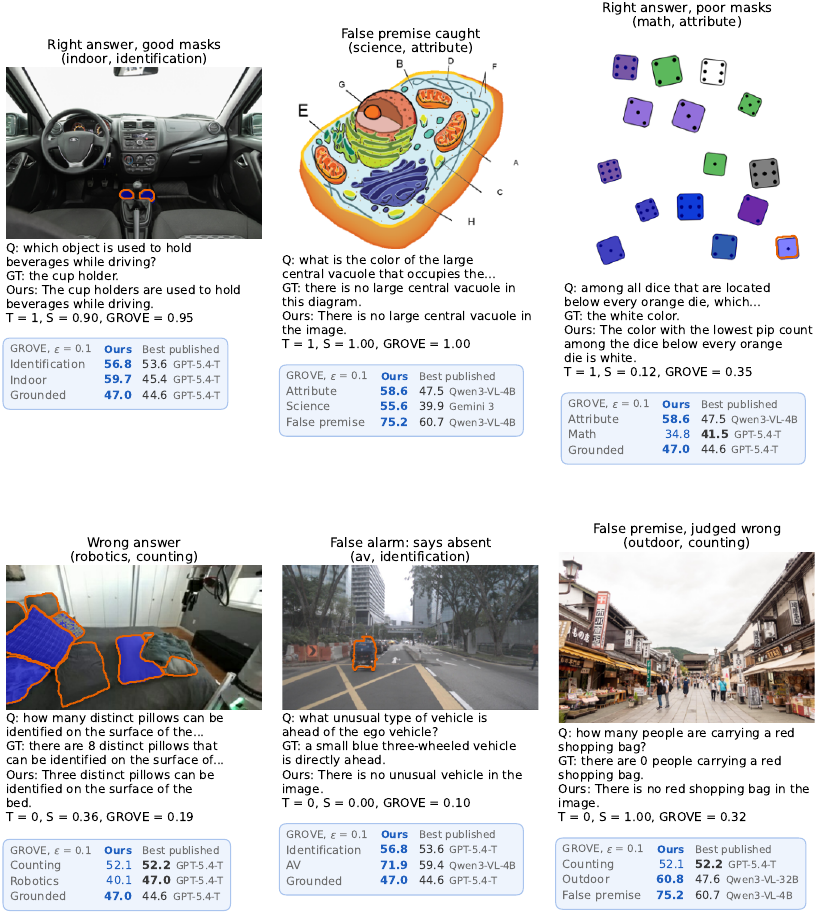}
\vspace{-3pt}%
\caption{The entry on six VistaQA samples. Orange outline: ground-truth masks; blue fill: our masks. $T$: judge verdict
(AWQ); $S$: mask score\ishraq{; every GROVE value in the figure uses $\epsilon=0.1$}. Blue boxes: our entry's GROVE (AWQ
judge) beside the best published system's on the panel's task, domain and question type
(Table~\ref{tab:main}), with the higher value in bold. Published per-sample outputs are not
released, so the comparison is by category rather than by sample. The wrong answers include undercounting and premise false
alarms on small objects; in the last panel, the judge rejects an answer that does reject the false premise. Images: VistaQA
(CC BY 4.0).}
\label{fig:qual}
\vspace{-6pt}%
\end{figure}

\ishraq{Figure~\ref{fig:qual} shows the entry on six VistaQA samples, one per outcome. We drew the panels after development
closed, picking each sample with a fixed seed from all qualifying samples and each from a different domain. The selection
rule was registered before any sample was viewed. Further image inspection, including selection for Figure~\ref{fig:pipe},
was solely for post-development illustration and did not inform system design.}

\clearpage
\section{\texorpdfstring{\ishraq{Error analysis of the entry}}{Error analysis of the entry}}\label{app:anat}
\ishraq{We analyze how the entry’s score is distributed across its predictions, identifying the main sources of gains and losses. All numbers aggregate the entry's
predictions and the official evaluator's per-sample output; they were computed after the entry was frozen and scored, so
they describe the entry and changed nothing in it. We recover the per-sample mask score $S$ from the per-sample GROVE $g$
and the verdict ($S=g^2$ if $T=1$, $S=(g/0.1)^2$ if $T=0$, and $S=0$ at the floor), and the mean of the recovered values
reproduces the official mIoU of 52.05.}

\begin{figure}[!htbp]
\color{ishraq}
\centering
\includegraphics[width=\textwidth]{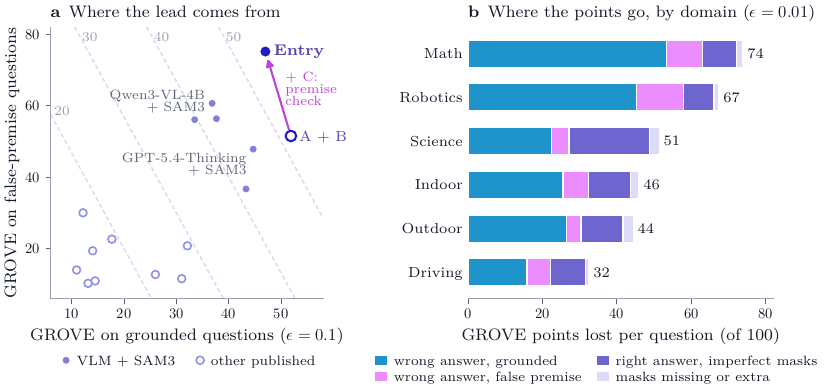}
\caption{The entry at a glance (AWQ judge). (a) GROVE on grounded questions against GROVE on false-premise questions for the
14 published systems and for ours, all at $\epsilon=0.1$; each diagonal joins points of equal overall GROVE, and the arrow
is the premise check. (b) GROVE points lost per question of each domain at $\epsilon=0.01$, split by cause.}
\label{fig:entry}
\end{figure}

\ishraq{\textbf{Where the lead comes from} (Figure~\ref{fig:entry}a, at the benchmark paper's $\epsilon=0.1$). Answers and
boxes alone (A + B) already lie ahead of GPT-5.4-Thinking + SAM3 on both axes, at 51.9 on grounded and 51.6 on
false-premise questions against 44.7 and 47.9. The premise check then gives up 4.2 points on grounded questions to gain
23.6 on false-premise ones. Because 27\% of VistaQA's questions rest on a false premise, the trade is worth 3.3 points
overall, and it moves most of the entry's margin onto false-premise questions.}

\ishraq{\textbf{Task by domain} (Table~\ref{tab:td}). Driving scenes are the best domain for five of the six tasks, and math
or robotics is the weakest for every task. The best cell is counting in driving scenes (79.9), and the worst is comparison
in math (13.8).}

\begin{table}[!htbp]
\centering\small
\caption{The entry's GROVE (AWQ judge) per task and domain ($n$ = 21--46 per cell; robotics has no OCR questions). In each
row, bold marks the best domain and \textcolor{loss}{color} the weakest.}
\label{tab:td}
\vspace{-2pt}%
\begin{tabular}{lrrrrrrr}
\toprule
 & \multicolumn{6}{c}{\textbf{Domain}} & \\
\cmidrule(lr){2-7}
\textbf{Task} & \textbf{AV} & \textbf{Indoor} & \textbf{Outdoor} & \textbf{Science} & \textbf{Robotics} & \textbf{Math} & \textbf{All} \\
\midrule
Identification & \textbf{66.4} & 60.7 & 61.9 & 44.2 & 41.0 & \textcolor{loss}{23.3} & 50.8 \\
Attribute & 64.9 & 58.5 & 59.8 & \textbf{68.0} & \textcolor{loss}{31.1} & 34.9 & 52.6 \\
OCR & \textbf{75.0} & 59.9 & 67.7 & 54.3 & -- & \textcolor{loss}{19.2} & 58.4 \\
Spatial & \textbf{61.7} & 42.7 & 56.0 & 38.0 & 25.9 & \textcolor{loss}{23.1} & 41.8 \\
Counting & \textbf{79.9} & 48.8 & 41.5 & 39.8 & \textcolor{loss}{31.2} & 41.8 & 44.6 \\
Comparison & \textbf{60.2} & 51.6 & 46.0 & 47.5 & 33.0 & \textcolor{loss}{13.8} & 43.2 \\
\midrule
\ours All & \textbf{67.5} & 54.1 & 55.6 & 48.6 & 32.6 & \textcolor{loss}{26.1} & 48.4 \\
\bottomrule
\end{tabular}
\vspace{-6pt}%
\end{table}

\ishraq{\textbf{Where the points go} (Table~\ref{tab:loss}, Figure~\ref{fig:entry}b). Wrong answers on grounded questions cost
30.43 of the 51.60 points lost (59\%). They are the largest item in every group, though only narrowly in science (3.95,
against 3.82 for imperfect masks), and robotics loses the most in total. Counting loses little to masks (0.61) because its
right answers are few and have good masks (mean $S$ of 73.8); comparison is the reverse (2.98, with a mean $S$ of 36.7).
Per question, math and robotics lose 74 and 67 of 100 points, mostly to wrong answers, whereas science loses almost as much
to imperfect masks as to wrong answers.}

\begin{table}[!htbp]
\centering\scriptsize
\setlength{\tabcolsep}{4pt}
\caption{Decomposition of GROVE score losses by task, domain, and instance count. For each group, we report the number of samples, GROVE, text accuracy, and mean mask score ($S$) on correctly answered grounded questions. The remaining columns break down the GROVE points lost by error type, expressed as contributions to the overall 51.60-point loss. Grounded-question losses comprise wrong answers, imperfect masks, and missing masks; false-premise losses comprise incorrect answers and unnecessary masks. The instance-count breakdown applies to grounded questions only.}
\label{tab:loss}
\vspace{-2pt}%
\begin{tabular}{lrrrrrrrrrr}
\toprule
 & & & & \textbf{Mean} & \multicolumn{3}{c}{\textbf{Lost: grounded}} & \multicolumn{2}{c}{\textbf{Lost: false premise}} & \textbf{Total} \\
\cmidrule(lr){6-8}\cmidrule(lr){9-10}
\textbf{Group} & $n$ & \textbf{GROVE} & \textbf{Text} & $S$ & \textbf{Wrong} & \textbf{Imperfect} & \textbf{No mask} & \textbf{Wrong} & \textbf{Masks} & \textbf{lost} \\
\midrule
Identification & 214 & 50.8 & 60.3 & 66.7 & 5.52 & 1.78 & 0.23 & 1.57 & 0.00 & 9.11 \\
Attribute & 191 & 52.6 & 64.4 & 56.8 & 4.66 & 1.71 & 0.47 & 0.99 & 0.00 & 7.82 \\
OCR & 173 & 58.4 & 75.7 & 54.0 & 3.22 & 2.60 & 0.00 & 0.25 & 0.16 & 6.22 \\
Spatial & 199 & 41.8 & 56.3 & 46.9 & 5.43 & 2.27 & 0.16 & 1.85 & 0.31 & 10.02 \\
Counting & 188 & 44.6 & 47.9 & 73.8 & 7.02 & 0.61 & 0.23 & 1.07 & 0.08 & 9.01 \\
Comparison & 192 & 43.2 & 61.5 & 36.7 & 4.57 & 2.98 & 0.23 & 1.56 & 0.08 & 9.42 \\
\midrule
AV & 202 & 67.5 & 76.2 & 70.1 & 2.76 & 1.64 & 0.08 & 1.11 & 0.08 & 5.67 \\
Indoor & 205 & 54.1 & 65.9 & 58.3 & 4.52 & 2.00 & 0.23 & 1.22 & 0.16 & 8.13 \\
Outdoor & 196 & 55.6 & 68.4 & 58.2 & 4.46 & 1.91 & 0.39 & 0.68 & 0.08 & 7.51 \\
Science & 204 & 48.6 & 72.1 & 38.5 & 3.95 & 3.82 & 0.23 & 0.82 & 0.23 & 9.06 \\
Robotics & 200 & 32.6 & 40.5 & 51.5 & 7.84 & 1.41 & 0.23 & 2.17 & 0.00 & 11.65 \\
Math & 150 & 26.1 & 34.7 & 42.3 & 6.90 & 1.16 & 0.16 & 1.28 & 0.08 & 9.58 \\
\midrule
Single-instance & 596 & 43.1 & 62.8 & 53.1 & 18.50 & 9.80 & 1.01 & 0.00 & 0.00 & 29.32 \\
Multi-instance & 247 & 32.7 & 41.7 & 61.4 & 11.93 & 2.14 & 0.31 & 0.00 & 0.00 & 14.38 \\
\midrule
\ours All & 1157 & 48.4 & 60.8 & 54.9 & 30.43 & 11.94 & 1.32 & 7.28 & 0.62 & 51.60 \\
\bottomrule
\end{tabular}
\vspace{-6pt}%
\end{table}

\begin{figure}[!htbp]
\color{ishraq}
\centering
\includegraphics[width=\textwidth]{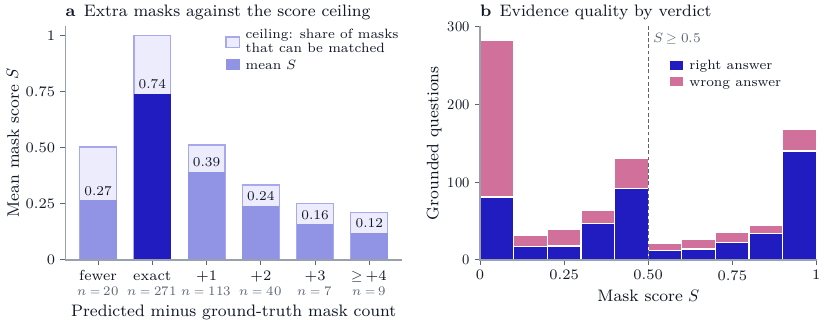}
\caption{Masks of the entry on grounded questions. (a) Mean $S$ of the 460 right answers with masks, by predicted minus
ground-truth mask count; the pale bar behind each is the group's ceiling, the share of masks that can be matched at all. (b)
Distribution of $S$ over the 843 grounded questions, by the judge's verdict.}
\label{fig:masks}
\end{figure}

\ishraq{\textbf{Answer against evidence} (Figure~\ref{fig:masks}b). On the 843 grounded questions, the entry gives a right
answer with good evidence ($S\geq0.5$) on 26.3\% and a wrong answer with weak evidence on 34.6\%. On 8.8\%, the evidence is
good but the answer wrong, and on 30.2\%, the answer is right but the evidence weak. The distribution of $S$ has two modes:
of the 366 wrong answers, 201 have an $S$ below 0.1, and of the 477 right ones, 140 have an $S$ of at least 0.9, so the
system mostly either finds the evidence or misses it altogether. Compared with answers and boxes alone (A + B), the premise
check and the count raise move a net 63 samples into the wrong-and-weak group.}

\ishraq{\textbf{Extra masks cost through the count, not through mask quality} (Figure~\ref{fig:masks}a). The entry predicts no
mask for 383 samples, and 314 samples have none in the ground truth. Of the 460 right grounded answers with masks, 271 have
the exact number (mean $S$ of 0.74), 169 too many (0.33) and 20 too few (0.27). The excess sits on single-instance
questions: of those 361 answers, 200 are exact and 161 have too many, whereas of the 99 multi-instance ones, 71 are exact,
8 have too many and 20 too few. $S$ divides the matched IoU by the larger of the two mask counts, so an answer with one
extra mask on a single-instance question can reach 0.5 at most. Measured against this ceiling, answers with one or two
extra masks reach 77\% and 73\% of it (0.39 of 0.51 and 0.24 of 0.33), as exact answers do (74\%). The masks that match
are thus as good as elsewhere, and the loss comes from the count alone. If the 169 over-masked answers scored like the
exact ones, GROVE would rise by about 4.0 points, which is the headroom that Appendix~\ref{app:mv} goes after.}

\begin{figure}[!htbp]
\color{ishraq}
\centering
\includegraphics[width=\textwidth]{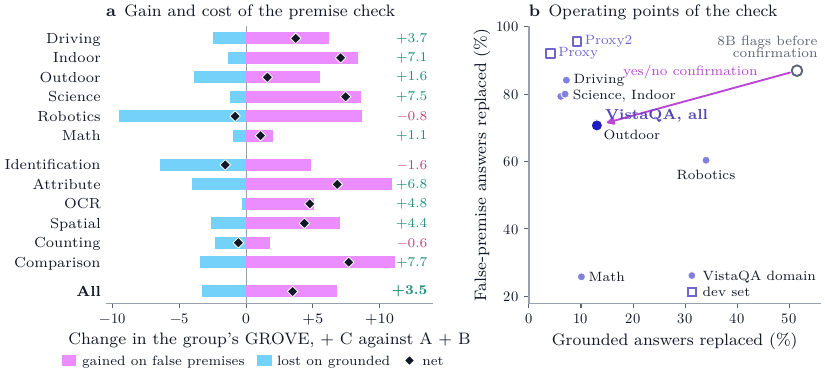}
\caption{The premise check on VistaQA, + C against A + B. (a) Change in each group's GROVE: points gained on false-premise
questions, points lost on grounded ones, and the net. (b) Share of false-premise answers that the check replaces against
the share of grounded answers that it replaces (false alarms), per domain, with the two dev sets and the 8B's flags before
the yes/no confirmation.}
\label{fig:premise}
\end{figure}

\ishraq{\textbf{Premise check} (Table~\ref{tab:prem}, Figure~\ref{fig:premise}). The 8B flags 707 of the 1,157 questions: 273
of the 314 false-premise questions and 434 of the 843 grounded ones. The yes/no confirmation keeps 335 of these flags (225
and 110), giving up 48 hits to remove 324 false alarms, and the answer text changes on 222 of the 225. Without the check,
the 27B already denies the premise in 96 false-premise and 7 grounded answers. The check gains on false-premise questions
in every group and loses on grounded ones. Its net effect is negative for robotics, where it replaces 34\% of grounded
answers, and for identification and counting. Of the 110 grounded answers it replaced, 63 had been right.}

\ishraq{Figure~\ref{fig:premise}b shows the same check as one operating point per domain. Driving, indoor and science
questions sit together, with 79\% to 84\% of false-premise answers replaced at 6\% to 7\% false alarms. Robotics pays 34\%
false alarms for 60\% of its false premises, and math catches only 26\% of them. Both dev sets lie above every VistaQA
domain (92\% and 96\% of false premises caught), so they overstated the check's recall; Appendix~\ref{app:transfer} returns
to this gap.}

\begin{table}[!htbp]
\centering\scriptsize
\caption{Premise check of the entry per group, against A + B: the answers it changed among false-premise questions and among
grounded ones (false alarms), how many of the latter had been right, and the change in the group's GROVE from each side. The
net change is colored by its sign.}
\label{tab:prem}
\vspace{-2pt}%
\begin{tabular}{lrrrrrr}
\toprule
\textbf{Group} & \textbf{False premise changed} & \textbf{Grounded changed} & \textbf{Were right} & $\Delta$\,\textbf{false premise}
 & $\Delta$\,\textbf{grounded} & \textbf{Net} \\
\midrule
AV & 53\,/\,63\pct{84} & 10\,/\,139\pct{7} & 6 & $+6.24$ & $-2.52$ & \textcolor{gain}{$+3.72$} \\
Indoor & 46\,/\,58\pct{79} & 9\,/\,147\pct{6} & 5 & $+8.43$ & $-1.34$ & \textcolor{gain}{$+7.09$} \\
Outdoor & 35\,/\,49\pct{71} & 19\,/\,147\pct{13} & 11 & $+5.51$ & $-3.92$ & \textcolor{gain}{$+1.59$} \\
Science & 48\,/\,60\pct{80} & 10\,/\,144\pct{7} & 7 & $+8.65$ & $-1.18$ & \textcolor{gain}{$+7.47$} \\
Robotics & 32\,/\,53\pct{60} & 50\,/\,147\pct{34} & 29 & $+8.73$ & $-9.56$ & \textcolor{loss}{$-0.83$} \\
Math & 8\,/\,31\pct{26} & 12\,/\,119\pct{10} & 5 & $+2.04$ & $-0.96$ & \textcolor{gain}{$+1.08$} \\
\midrule
Identification & 37\,/\,52\pct{71} & 29\,/\,162\pct{18} & 19 & $+4.88$ & $-6.44$ & \textcolor{loss}{$-1.56$} \\
Attribute & 43\,/\,55\pct{78} & 23\,/\,136\pct{17} & 14 & $+10.93$ & $-4.09$ & \textcolor{gain}{$+6.84$} \\
OCR & 30\,/\,45\pct{67} & 3\,/\,128\pct{2} & 1 & $+5.10$ & $-0.32$ & \textcolor{gain}{$+4.78$} \\
Spatial & 33\,/\,59\pct{56} & 24\,/\,140\pct{17} & 11 & $+7.01$ & $-2.64$ & \textcolor{gain}{$+4.37$} \\
Counting & 39\,/\,52\pct{75} & 15\,/\,136\pct{11} & 7 & $+1.77$ & $-2.35$ & \textcolor{loss}{$-0.58$} \\
Comparison & 40\,/\,51\pct{78} & 16\,/\,141\pct{11} & 11 & $+11.16$ & $-3.46$ & \textcolor{gain}{$+7.70$} \\
\midrule
\ours All & 222\,/\,314\pct{71} & 110\,/\,843\pct{13} & 63 & $+6.81$ & $-3.32$ & \textcolor{gain}{$+3.49$} \\
\bottomrule
\end{tabular}
\vspace{-6pt}%
\end{table}

\ishraq{\textbf{Count raise.} Of the 238 questions that match the count regular expression, 97 pass the plain-count guard.
The rule raised 23 answers: 10 went from right to wrong, 2 from wrong to right, and 11 were wrong before and after. Eleven
of the 23 lost their masks through the bug noted in Section~\ref{sec:results}.} The rule takes the winning counter's masks,
and on VistaQA the 8B count pass ran without masks (on the proxy it had them). With the masks kept, the same answers score
48.47 (a registered fix; Table~\ref{tab:transfer}), so the bug cost 0.07 and the wrongly raised counts the other 0.48.

\ishraq{\textbf{Image sizes} (image metadata only). The median image has 1.44\,MP and the largest 5.03\,MP. Of the 1,157
images, 536 exceed the answerer's 1.64\,MP cap and are downscaled for the VLM only. Median per domain: driving
1600$\times$900 (0 of 202 above the cap); indoor and outdoor 2250$\times$1500 (all 205 and 196); math 3.24\,MP (130 of
150); science 0.21\,MP (5 of 204); robotics 320$\times$180 (0 of 200). Robotics images are therefore very small, and we did
not test upscaling them.}

\section{\texorpdfstring{\ishraq{Judge sensitivity analysis}}{Judge sensitivity analysis}}\label{app:judge}
\ishraq{We examine how sensitive the reported scores are to the judge’s model build, its repeatability, and the evaluation harness.}

\begin{figure}[!htbp]
\color{ishraq}
\centering
\includegraphics[width=\textwidth]{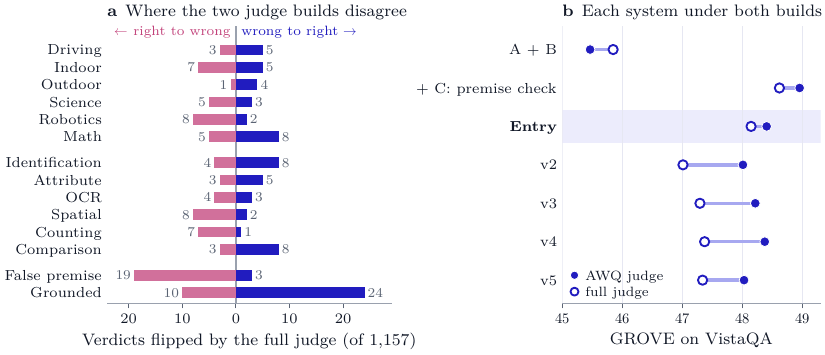}
\caption{The two builds of the judge on VistaQA. (a) Verdicts on the entry's unchanged predictions that flip when the full
judge replaces the AWQ build, per group. (b) GROVE of each system under both builds; v2 to v5 are the later systems of
Appendix~\ref{app:v2}.}
\label{fig:judge}
\end{figure}

\ishraq{\textbf{Two builds of the judge.} Re-judging the entry's unchanged predictions with the unquantized model flips 56 of
the 1,157 verdicts (4.8\%) and moves GROVE from 48.40 to 48.14 ($-0.26$, CI $-1.16$ to $+0.63$). Grounded questions gain
(24 verdicts up, 10 down), and false-premise questions lose (3 up, 19 down). Figure~\ref{fig:judge}a gives the flips per
group: the full judge is stricter on robotics, spatial and counting questions and more lenient on math, identification
and comparison. On the proxy development set, the two builds flip 8 of 560 verdicts (73.81 $\to$ 73.33); the full judge
is stricter on OCR casing and near-synonyms.}

\ishraq{\textbf{The ordering of the systems holds under both builds} (Figure~\ref{fig:judge}b). The premise-check stage and
the entry stay above every later system with either judge. Under the full judge, v2, v3, v4 and v5 trail the entry by 1.13, 0.85, 0.77 and 0.81, each with an interval
that includes zero (for v2, $-3.01$ to $+0.79$), so neither build separates them from the entry.}

\ishraq{\textbf{Repeatability and quirks.} The same judge build is not perfectly repeatable: across two runs, 3 of 802
identical VistaQA answers received a different verdict, which moved one analysis score by 0.06. The AWQ judge accepted 3 of
5 empty answers, and the full judge 2 of 5. These came from a first v5 run in which six large images failed (the premise
override filled one answer); the six fell back to the default input size before v5 was merged and judged. One
false-premise answer that does reject the premise was judged wrong (Appendix~\ref{app:qual}, last panel). The intervals
resample questions, so they do not include this judge noise.}

\ishraq{\textbf{Evaluation checks conducted before the first system was scored.} Submitting ground truth as a prediction yields a score of 100.00.
Moving each answer to the next sample while keeping the ground-truth masks gives 1.21\% text accuracy and a GROVE of 11.09,
exactly as predicted by the formula, $100\,(0.0121 + 0.9879 \times 0.1)$. Re-running SAM2.1 on stored boxes reproduces 71.74
on the proxy development set (548 of 560 predictions identical, the rest differing only numerically), and reusing stored
verdicts reproduces a judged run exactly (73.05).}

\ishraq{\textbf{Judge settings.} Both builds use the official prompt and parser, temperature 0 and at most 256 output
tokens. The full judge ran in fp16 with a 1,024-token context (2,560 for v5, whose longest answer has 1,844 tokens).}

\section{\texorpdfstring{\ishraq{Development sets and decisions}}{Development sets and decisions}}\label{app:dev}
\ishraq{Table~\ref{tab:sets} lists the dev sets and probes. The proxy development set and the premise, count and OCR probes
decided the entry (Table~\ref{tab:pre}), which reached 65.07 with answers and boxes, 72.27 with the premise check and 73.81
with the count raise. The other sets came after the entry was scored and served only the analysis systems
(Table~\ref{tab:post}). No number in these tables comes from VistaQA. In proxy2, the driving-scene masks are SAM3 masks of
annotated boxes, so those questions test which objects were found rather than mask quality.}

\begin{table}[!htbp]
\centering\scriptsize
\caption{Development datasets and evaluation probes. We report sample counts ($n$), data sources, false-premise questions (FP), mean masks per grounded sample, and task composition. VistaQA's mean mask count (2.06) is included for reference. Probes contain text ground truth only.}
\label{tab:sets}
\vspace{-2pt}%
\begin{tabular}{@{}>{\raggedright\arraybackslash}p{.72in}r>{\raggedright\arraybackslash}p{1.75in}rr>{\raggedright\arraybackslash}p{1.3in}@{}}
\toprule
\textbf{Set} & $n$ & \textbf{Source} & \textbf{FP} & \textbf{Masks} & \textbf{Composition} \\
\midrule
\ishraq{Proxy development set} & 560 & COCO val2017, PACO-LVIS, COCO-Text v2; templated questions, answers and masks from annotations & 150 & 1.80 & counting 100, identification 70, spatial 70, comparison 60, attribute 60, OCR 50 \\
Proxy2 & 580 & Drawn charts and geometry, BOP (YCB-V, LM-O), 2018 Data Science Bowl, BDD100K & 158 & 4.08 & spatial 120, counting 116, attribute 107, comparison 101, identification 87, OCR 49; math 150, robotics 150, AV 150, science 130 \\
Proxy3 & 200 & RefCOCOg validation (UMD split), crowded scenes & 0 & 1.65 & spatial 140, comparison 60 \\
\midrule
Premise probe & 300 & MathVista testmini: 200 questions on their own image, 100 moved to another question's image & 100 & -- & false-alarm and recall check on figures \\
Count probe & 168 & MathVista testmini free-form integer questions matching the count regex & 0 & -- & count guard \\
OCR probe & 225 & TextVQA val reading questions, top answer with at least 5 of 10 votes & 0 & -- & OCR zoom; premise false alarms \\
Math probe & 200 & MathVista testmini & 0 & -- & thinking, calculator \\
V*Bench probe & 238 & V*Bench (SA-1B photos, median 3.38\,MP), with target boxes & 0 & -- & input resolution \\
WSP practice & 300 images & COCO val2017 outside \ishraq{the proxy development set}; each image with one absent-object and one present-object question & half & -- & Appendix~\ref{app:wsp} only \\
\bottomrule
\end{tabular}
\vspace{-6pt}%
\end{table}

\begin{table}[!htbp]
\centering\scriptsize
\caption{Evaluation decisions and their measured outcomes on the proxy development set and targeted probes. Shaded rows summarize the final configuration choices.}
\label{tab:pre}
\vspace{-2pt}%
\begin{tabular}{>{\raggedright\arraybackslash}p{1.15in}>{\raggedright\arraybackslash}p{3.23in}>{\raggedright\arraybackslash}p{.6in}}
\toprule
\textbf{Decision} & \textbf{Measured} & \textbf{Outcome} \\
\midrule
Premise on figures & Premise probe: 86 of 100 moved questions caught, 21 of 200 false alarms; OCR probe: 8 of 225 false alarms & No change \\
SAM3 veto of the premise check & $\tau=0.5$--$0.7$: 73.16--73.18 \vs 73.31 & Dropped \\
OCR zoom re-read & \ishraq{Proxy development set} 73.96 (one net sample); OCR probe text 76.44 $\to$ 75.11 & Dropped \\
Thinking in the 27B & 73.64 \vs 73.81 at three times the inference time; 34 of 560 traces unclosed & Dropped \\
Offline combinations & Concept masks on counting questions +0.05 to +0.16 (better and worse about equally often). Requiring the answers to agree with the premise check loses 43 hits to save 16 false alarms. Thinking against no-think: both right 432, both wrong 97, only one right 15 and 16, so no signal for a selector & None adopted \\
\midrule
\ours Answerer (one pass, SAM2.1-L) & Qwen3-VL-4B 39.02 / 55.7; Qwen3-VL-8B 58.60 / 66.4; Qwen3.5-9B 62.14 / 74.5; Qwen3.6-27B 64.91 / 71.2; Qwen3.6-35B-A3B not run (its fused experts stay unquantized in 4-bit loading) & 27B \\
\ours Two stages (reason, then ground each target) & 8B 58.56 / 65.0 against one pass 58.60 / 66.4; with an existence check 57.41 / 64.8; 4B two stages with the check 50.20 / 58.0 & One pass \\
\ours Short-answer rewrite & 8B 58.60 $\to$ 62.31 (296 of 560 answers had at most 3 words); 27B chain 71.74 $\to$ 72.29 & Kept \\
\ours Segmenter & SAM2.1-L 71.74, with multimask 71.53; SAM3 71.72; filled boxes for OCR: $S$ 0.649 \vs SAM3 0.696 & SAM3 (also text-prompted) \\
\ours Premise checker & On the 27B's answers: 8B + confirm 71.74 (138 of 150 hits, 17 of 410 false alarms); the 27B's own check 69.60 (135, 36). 8B without confirm: 148, 105; reworded confirm: 133, 16; requiring both models: 73.39 \vs 73.31 & 8B + confirm \\
\ours Count raise & SAM3 only at $\tau=0.7$: 73.04; with the grammar fix 73.31; all three counters at $\tau=0.8$: 73.81 (exact on 72 of 98 counting questions; the 27B alone 60) & Kept \\
\ours Count guard & Count probe, exact counts of 168: guarded 115, unguarded 108, no raise 114; judge text 75.60, 70.83, 75.00 & Guard kept \\
\bottomrule
\end{tabular}
\vspace{-6pt}%
\end{table}

\begin{table}[!htbp]
\centering\scriptsize
\caption{Evaluation of additional system modifications on the development sets and targeted probes, with measured results and resulting configuration choices. }
\label{tab:post}
\vspace{-2pt}%
\begin{tabular}{@{}>{\raggedright\arraybackslash}p{1.15in}>{\raggedright\arraybackslash}p{3.4in}>{\raggedright\arraybackslash}p{.6in}@{}}
\toprule
\textbf{Test} & \textbf{Measured} & \textbf{Outcome} \\
\midrule
Qwen3.8-27B in the entry's chain & \ishraq{Proxy development set} 74.70 \vs 73.81 ($+0.89$, CI $-0.62$ to $+2.49$); proxy2 at the premise stage 61.04 \vs 57.53 & v2 \\
Gemma 4 31B answerer & \ishraq{Proxy development set} 69.77, text 79.5, mIoU 70.7, OCR 25.3 (its boxes come as $[y_1,x_1,y_2,x_2]$, swapped before SAM3) & No \\
Three-model majority & 316 of 410 grounded answers right; Qwen3.8 alone 320, oracle 338 & No \\
Evidence prompt naming the answer element & Everywhere on proxy2: 57.64 \vs 57.42, but text 71.4 \vs 73.8; math rises 68.0 $\to$ 78.0 & Figures only \\
Figure router & Yes on \ishraq{the proxy development set} 0 of 560; proxy2 math 150 of 150, science 2 of 130, driving and robotics 0 of 150; MathVista 199 of 200; TextVQA 11 of 225 & v2 \\
SAM3 veto threshold $\tau$ & Pooled over \ishraq{the proxy development set} and proxy2: none 67.96; 0.7: 68.22; 0.6: 68.39; 0.5: 68.37; \textbf{0.4: 68.65}; 0.3: 68.45 & $\tau=0.4$, v2 \\
Premise AND with Qwen3.8 & Proxy2 hits 139 of 158, false alarms 28 of 422 (8B alone 151, 39) & No \\
Thinking on all routed figures & MathVista 122 $\to$ 142 of 200; proxy2 65.42, below 65.87 & No \\
Reasoning router + thinking & MathVista 143 of 200; proxy2 65.87 & v3 \\
Thinking budget 8,192 tokens & Closes 10 of 18 unclosed traces, one more right answer & No \\
Answer selector (Qwen3.8 picks, Qwen3.6 unless overruled twice) & \ishraq{Proxy development set} 65.46 (Qwen3.6 65.07, Qwen3.8 63.88, best-of-two ceiling 70.76); proxy2 56.85 (55.70, 60.04, 66.34); mostly position bias & No \\
Near-duplicate masks (IoU $\geq$ 0.5) & 11, 1, 1 and 0 masks dropped on four runs, changes of at most 0.01 & No \\
Three-way premise AND with veto & Proxy2 65.31 $\to$ 65.66; \ishraq{proxy development set} 72.47 $\to$ 72.07 & No \\
Calculator tool & MathVista of 200: no thinking 122; always think 142; v3 recipe 143; tool instead of thinking 135; thinking, tool where the trace ran out 146; always the tool 140; proxy2 unchanged & Fallback, v4 \\
4\,MP input & V*Bench text 176 $\to$ 192 of 238, box hit 194 $\to$ 209 of 292 & v5 \\
Mask clean-up & \ishraq{Proxy development set} / proxy2: hole filling $+0.43$ / $-0.05$; largest component $-1.30$ / $-0.16$; both $-0.85$ / $-0.21$ & No \\
\bottomrule
\end{tabular}
\vspace{-6pt}%
\end{table}

\section{\texorpdfstring{\ishraq{Post-freeze systems analysis (v2--v5)}}{Post-freeze systems analysis (v2--v5)}}\label{app:v2}
\ishraq{We analyze four subsequent system variants (v2–v5), examining how changes motivated by the observed VistaQA results affected performance across the evaluation sets. The entry's VistaQA aggregates
(premise false alarms on 13\% of grounded questions, over-masking, weak math and robotics) prompted a second system, built
without looking at any VistaQA sample. Because the proxy development set is COCO-only, we added proxy2: 580 questions (158
with a false premise) on drawn charts and geometry, BOP scenes~\cite{bop23} (YCB-V~\cite{posecnn}, LM-O~\cite{lmo}), 2018
Data Science Bowl nuclei~\cite{dsb2018} and BDD100K~\cite{bdd100k}, with answers and masks from annotations or drawing code
and none of VistaQA's image sources. Each system below was chosen on the dev sets and registered before its single VistaQA
run. Scores in parentheses use the full judge.}

\ishraq{\textbf{v2: four dev-set wins that did not carry over.} Each of v2's four changes won on the dev sets: (i)
Qwen3.8-27B~\cite{qwen38} answers; (ii) a Qwen3-VL-8B yes/no router sends charts, tables and diagrams to an evidence prompt
that boxes the answer element (the bar, the cell) instead of labels and legends; (iii) a SAM3 presence veto keeps the
answer when SAM3 finds the object that the premise check calls missing (score $\geq 0.4$, picked by GROVE pooled over both
dev sets); and (iv) the count raise keeps its masks. v2 gained 8.4 on proxy2 and 0.4 on the proxy development set but
scored 48.01 (47.01) on VistaQA, level with the entry ($-0.40$, 95\% CI $-2.38$ to $+1.55$; $-1.13$ with the full judge).
Stage by stage, the new backbone, +1.7 on proxy2, lost 3.0 on VistaQA at the answer stage (42.43 \vs 45.46, on
false-premise, math and science questions), and the figure prompt, +2.6 on proxy2's drawn charts, gave nothing. With the
veto, ``not present'' answers fell from 110 to 64 on grounded questions but also from 225 to 189 on false-premise ones, a
small net gain (+0.2 on v2; added to the entry with the masks kept, it gives 49.03, computed \textit{post hoc} and not a
submission). Keeping only the first mask on non-counting answers, a registered test of the over-masking found in
Section~\ref{sec:results}, scored 47.18.}

\ishraq{\textbf{v3: thinking on multi-step questions.} Thinking on every routed figure raised MathVista probe accuracy from
61\% to 71\% but lowered proxy2, whose figures ask for values read off a chart. A second yes/no router (does the question
need several steps of reasoning?) kept the gain (143 of 200) and left proxy2 unchanged; registered before v2 was scored, it
became v3. On VistaQA, it sent 126 of the 353 figure questions to thinking: 22 answers turned right and 8 wrong (math 21.9
to 24.7), so v3 scored 48.21 (47.29), $+0.21$ over v2 (95\% CI $-0.35$ to $+0.75$) and level with the entry.}

\ishraq{\textbf{v4: a calculator tool.} A larger thinking budget is no fix: 27 of v3's traces hit 4,096 tokens, and on the
probe, a budget of 8,192 closed 10 of 18 such traces but gained only one answer. A calculator did better there: Qwen3.8
writes a short Python program that a sandbox runs (Appendix~\ref{app:prompts}), which raised the probe from 143 to 146 of
200. As v4, it ran on 21 of the 27 questions and changed 17 answers, gaining 3 right answers and losing 3, for 48.37
(47.37), $+0.16$ over v3 (95\% CI $-0.23$ to $+0.56$), all of it from judge flips on unchanged answers.}

\ishraq{\textbf{v5: larger input images.} Of the VistaQA images, 536 exceed the answerer's 1.64\,MP cap (no dev image does),
and on V*Bench~\cite{vstar} (238 questions on 3.4\,MP photos), a 4.19\,MP cap gained 16 answers. As v5, it changed 156
answers on VistaQA's large images, gaining 22 right answers and losing 23, for 48.03 (47.34), level with v4.}

\ishraq{The dev sets thus predicted the premise check's large gain, but not the effect of the backbone or of the figure
prompt; Appendix~\ref{app:transfer} compares dev-set and VistaQA effects for every change.}

\begin{figure}[!htbp]
\color{ishraq}
\centering
\includegraphics[width=\textwidth]{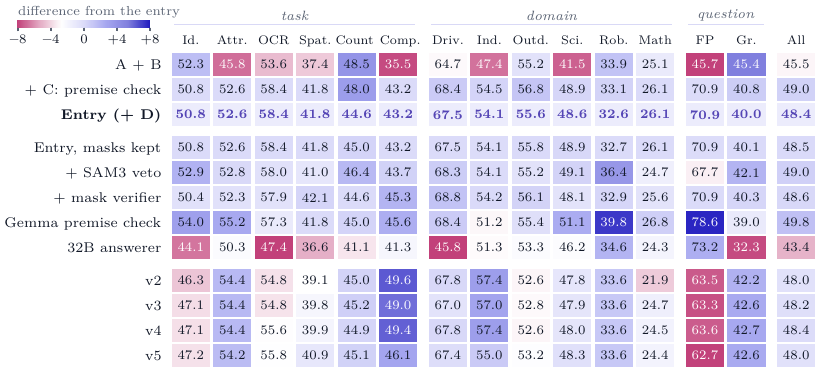}
\caption{GROVE (AWQ judge) of every system scored on VistaQA in Appendices~\ref{app:v2}, \ref{app:mv} and~\ref{app:ms}, per
task, domain and question type. Each cell is colored by its difference from the entry (bold, shaded row): raspberry where
lower, lavender to indigo where higher, clipped at $\pm 8$. The veto row adds the
SAM3 veto to the entry with the masks kept and was computed \textit{post hoc}.}
\label{fig:matrix}
\end{figure}

\ishraq{\textbf{Where each system gains and loses} (Figure~\ref{fig:matrix}). Broken down by category, v2 to v5 share one
profile. Against the entry, they gain on comparison and on grounded questions (+6.3 and +2.2 for v2) and lose 7.3 to 8.2
points on false-premise questions and 3.6 to 4.5 on identification, and the two sides cancel. The veto makes the same trade
on a smaller scale, with +3.8 on robotics and +2.0 on grounded questions for $-3.2$ on false-premise ones. The Gemma
premise check of Appendix~\ref{app:ms} moves the other way, +7.8 on false-premise questions and +7.2 on robotics for $-1.0$
on grounded ones, and the 32B answerer's loss is concentrated in driving scenes ($-21.8$) and OCR ($-10.9$). Across the
rows, no system gains on both question types at once: each gain on one side is matched by a loss, or by no change, on the
other.}

\section{\texorpdfstring{\ishraq{Weight-space programming analysis}}{Weight-space programming analysis}}\label{app:wsp}
\vspace{-2pt}%
\ishraq{After development closed, we conducted an exploratory study of weight-space programming~\cite{takehi2026wsp}, a
gradient-free approach in which a large language model inspects a smaller network's weights and activations and directly
proposes parameter modifications. Claude Opus 5.5 (Anthropic) edited Qwen3-VL-8B-Instruct. The intervention targeted the
model's single-pass handling of false-premise questions, which we measured on the proxy development set
(Section~\ref{sec:setup}, Table~\ref{tab:sets}). We built this set of 560 questions by filling fixed templates from the
public annotations of COCO val2017~\cite{coco}, PACO-LVIS~\cite{paco} and COCO-Text v2~\cite{cocotext}, so that every answer
and mask is computed from an annotation rather than written by a person or a model. Its 150 false-premise questions each ask
about the category that most often co-occurs with the objects in the image but is absent from it. On this set, the
unmodified model rejects only 68 of 150 absent objects, whereas its dedicated premise check (Pass~C) catches 138.}

\ishraq{To isolate the editing process from evaluation data, we built a practice set of 300 COCO val2017 images disjoint
from the proxy development set, each paired with a question about an absent, frequently co-occurring category and the same
question about a present object. We reserved 200 images for representation analysis and 100 for scoring candidates. At the first answer
token, the residual-stream direction separating absent-object questions from their present-object counterparts
distinguishes 97--99\% of pairs from layer 16 onward. This separation alone does not establish that the model rejects the
premise: the direction predicts its denials only from layer 20 onward (AUC 0.72 at layer 20, 0.87 at layer 24).}

\ishraq{Among 21 scored candidates, of a registered maximum of 40, we selected an intervention that combines two changes
(Table~\ref{tab:wsp-cand}). Rank-1 updates to the six output projections of layers 18--20, centered by a bias term, double
the contribution of the layer-21 direction, and a bias at layer 20 shifts the residual 0.3 class gaps, the distance between
the absent and present class means, toward the absent-object direction. On the held-out scoring images, this raised
absent-object denials from 75 to 85 of 100, while denials of present objects rose only from 4 to 5 of 100. Either change
alone gained about half as much, and a fivefold stronger bias reached 91 denials only at 13 false alarms. The selection
criteria, the scoring procedure and the paired VistaQA runs were registered before the edit was finalized, and the edit was
fixed before any development-set evaluation, which keeps selecting the intervention separate from assessing it.}

\begin{table}[!htbp]
\color{ishraq}
\centering\footnotesize
\setlength{\tabcolsep}{5pt}
\caption{Candidate edits on the held-out practice images: denials out of 100 absent and 100 present objects. Score is the
registered selection rule, the absent rate minus 1.35 times the present rate. Bold: best score.}
\label{tab:wsp-cand}
\vspace{-2pt}%
\begin{tabular}{lrrr}
\toprule
\textbf{Edit} & \textbf{Absent denied} $\uparrow$ & \textbf{Present denied} $\downarrow$ & \textbf{Score} $\uparrow$ \\
\midrule
None & 75 & 4 & 0.696 \\
Bias only, 0.3 class gaps & 81 & 5 & 0.742 \\
Bias only, 1.5 class gaps & 91 & 13 & 0.734 \\
Amplification only, layers 18--20 & 80 & 4 & 0.746 \\
Bias (0.3) + amplification, chosen & 85 & 5 & \textbf{0.782} \\
\bottomrule
\end{tabular}
\vspace{-6pt}%
\end{table}

\begin{table}[!htbp]
\color{ishraq}
\centering\footnotesize
\setlength{\tabcolsep}{5pt}
\caption{Weight edit against the premise check on Qwen3-VL-8B (AWQ judge, SAM3 box masks). All columns are GROVE except
Text, the text accuracy; FP and Gr. are false-premise and grounded questions. Bold: best per column.}
\label{tab:wsp}
\vspace{-2pt}%
\begin{tabular}{lrrrrrrr}
\toprule
 & \multicolumn{3}{c}{\textbf{Development sets}} & \multicolumn{4}{c}{\textbf{VistaQA}} \\
\cmidrule(lr){2-4}\cmidrule(l){5-8}
\textbf{System} & \textbf{Proxy} & \textbf{Proxy2} & \textbf{Pooled} & \textbf{GROVE} & \textbf{Text} & \textbf{FP} & \textbf{Gr.} \\
\midrule
8B, single pass & 58.42 & 48.35 & 53.30 & 38.76 & 49.8 & 39.7 & 38.4 \\
+ weight edit & 61.66 & 50.27 & 55.86 & 40.46 & 51.4 & 45.5 & \textbf{38.6} \\
+ premise check & 67.58 & 52.60 & 59.96 & 44.35 & 55.1 & 68.2 & 35.5 \\
+ weight edit + premise check & \textbf{68.01} & \textbf{52.73} & \textbf{60.24} & \textbf{44.90} & \textbf{55.4} & \textbf{69.8} & 35.6 \\
\bottomrule
\end{tabular}
\vspace{-6pt}%
\end{table}

\vspace{-1.5pt}%
\ishraq{\textbf{Results.} Table~\ref{tab:wsp} compares the edit with the premise check, with SAM3 box masks in every row.
The edited 8B improved single-pass GROVE by 2.57 on the proxy development set and proxy2 pooled (95\% CI $+1.71$ to $+3.50$)
and by 1.70 on VistaQA (95\% CI $+0.83$ to $+2.60$), where the intervention altered 154 answers, yielding 27 additional correct verdicts
against 8 lost. The improvement is concentrated in false-premise handling: false-premise GROVE rose from 39.7 to 45.5, while
grounded GROVE remained effectively unchanged. Overall, the edit recovers about 30\% of the improvement that the premise
check provides on VistaQA (1.70 of 5.59 points), at no additional inference-time computation and without the check's loss on
grounded questions. Applied alongside the premise check, however, the intervention provides no statistically significant
benefit ($+0.56$, 95\% CI $-0.02$ to $+1.18$). The entry takes its answers from Qwen3.6-27B, so this exploratory edit of the
8B does not alter the submitted system.}

\section{\texorpdfstring{\ishraq{Mask verifier analysis}}{Mask verifier analysis}}\label{app:mv}
\ishraq{After development ended, we tested a training-free approach to address the over-masking observed in Section~\ref{sec:results}. For each prediction with at least two
masks on a non-counting question, a Set-of-Mark check~\cite{yang2023som} draws each mask's bounding box with its number on
the image and asks a VLM which numbered boxes mark ``visual evidence required to support this answer'' (VistaQA's
definition), with the question and the answer in the prompt (Appendix~\ref{app:prompts}) and a JSON list as the reply.
Masks that it does not list are dropped, and without a valid list all are kept. Answers never change, so the text verdicts
are reused. We compared two checkers, the answerer (Qwen3.6-27B) and Qwen3-VL-8B.}

\ishraq{\textbf{A dev set with multiple objects per image.} The proxy development set and proxy2 hold few over-masked answers, so we built
\textit{proxy3}: 200 questions on RefCOCOg~\cite{mao2016refcocog} validation images (UMD
split~\cite{nagaraja2016refcocog}), each holding several objects of one category. Each question names its objects by a
referring expression: which side of the image, left or right of another object, or which appears larger. On proxy3, the
entry's answer and mask passes over-mask 19 of their 179 right grounded answers.}

\ishraq{\textbf{Result.} On the three dev sets pooled (1,340 questions), the 27B checker beat the 8B, which lost 0.65, but
still scored below no check at all (66.96 against 67.34; $-0.38$, CI $-0.63$ to $-0.15$). It fixed 15 of 49 over-masked
answers, but 32 of the 72 predictions it changed already had the right number of masks; in 26 of these, it kept only one of
the two objects in ``Which appears larger, the A or the B?'' or ``Is the A left or right of the B?''. On VistaQA, run as
registered, it dropped 111 masks from 98 predictions. Over-masked right answers fell from 169 to 140, 8 drops hit a correct
count, and mIoU rose from 52.05 to 52.50. GROVE moved from 48.40 to 48.61 ($+0.21$, CI $-0.13$ to $+0.53$).}

\ishraq{\textbf{Why the dev sets predicted a loss.} Over-masking is rare on the dev sets (6\% of right grounded answers with
masks, against 37\% on VistaQA), so a check that sometimes removes a needed mask had little to gain there. On VistaQA the
the potential gain is about 4.0 GROVE points (Appendix~\ref{app:anat}, Figure~\ref{fig:masks}a), and the check recovered 0.21
of them.}

\section{\texorpdfstring{\ishraq{Model swap analysis}}{Model swap analysis}}\label{app:ms}
\ishraq{After development closed, we tested two model swaps in the
entry's chain, each alone and registered before any of their results: Qwen3-VL-32B-Instruct~\cite{qwen3vl} as the answerer
in place of Qwen3.6-27B, and Gemma~4 31B (instruction-tuned) as the premise checker in place of Qwen3-VL-8B. Both ran in
4-bit like the 27B, with the entry's prompts, and every other pass and rule is unchanged. Proxy3 (Appendix~\ref{app:mv})
has no false premises, so every premise override there is a false alarm.}

\ishraq{\textbf{On the dev sets} (1,340 questions pooled), the entry's chain scores 66.27; it includes the premise check on
proxy3, so this differs from the mask verifier's base. The 32B answerer was level with it at 66.32 ($+0.05$, CI $-1.27$ to
$+1.42$), losing 2.0 on the proxy development set (COCO) and gaining 1.4 on proxy2 and 1.9 on proxy3. Gemma turned about
as many false-premise answers into ``not present'' as the 8B (105 \vs 106 of 150 on the proxy development set, 117 \vs 121
of 158 on proxy2) but raised more false alarms on grounded questions (35 \vs 17 of 410 on the proxy development set and 31
\vs 22 of 200 on proxy3, though 33 \vs 39 of 422 on proxy2), and it scored 64.47 ($-1.80$, CI $-3.11$ to $-0.50$).}

\ishraq{\textbf{On VistaQA}, run as registered, the ranking reversed. Gemma changed 239 of the 314 false-premise answers (the
8B: 222) and 120 of the 843 grounded ones (the 8B: 110). False-premise GROVE rose from 70.9 to 78.6, and grounded GROVE fell
from 40.0 to 39.0, for 49.76 overall, a gain of 1.35 (CI $-0.20$ to $+2.93$) that is not significant. The 32B scored 43.38,
a loss of 5.02 (CI $-6.92$ to $-3.15$). Its text accuracy fell from 60.8 to 58.4 and its mIoU from 52.1 to 47.3, most of
all on driving scenes, where GROVE fell from 67.5 to 45.8 while text accuracy fell only from 76.2 to 73.8
(Figure~\ref{fig:matrix}). Proxy2's BDD100K questions did not show this.}

\ishraq{\textbf{The driving-scene loss is in the masks.} Mean $S$ on right grounded answers fell from 70.1 to 26.2 on driving
scenes and moved by at most 2.3 in the other domains, with similar mask counts; driving scenes account for about 3.8 of the
5.02 points lost. On their 1600$\times$900 images, the 32B's single boxes have median ratios to the 27B's of 0.95
(x-center), 0.97 (y-center), 0.95 (width) and 0.88 (height), with $n=76$. In the other domains, the ratios are about 1.00,
and on proxy2's 1280$\times$720 BDD100K images, the center ratios are 1.00 ($n=54$). At $\epsilon=0.1$, published
Qwen3-VL-32B + SAM3 scores 58.40 on driving scenes where our run scores 54.0, so this looks like a coordinate shift at that
image size in our 4-bit run rather than a limit of the model. Our registered rule allows re-runs only for infrastructure
failures, so we did not re-run it.}

\section{\texorpdfstring{\ishraq{Transfer analysis: from dev sets to VistaQA}}{Transfer analysis: from dev sets to VistaQA}}\label{app:transfer}
\begin{figure}[!htbp]
\color{ishraq}
\centering
\includegraphics[width=\textwidth]{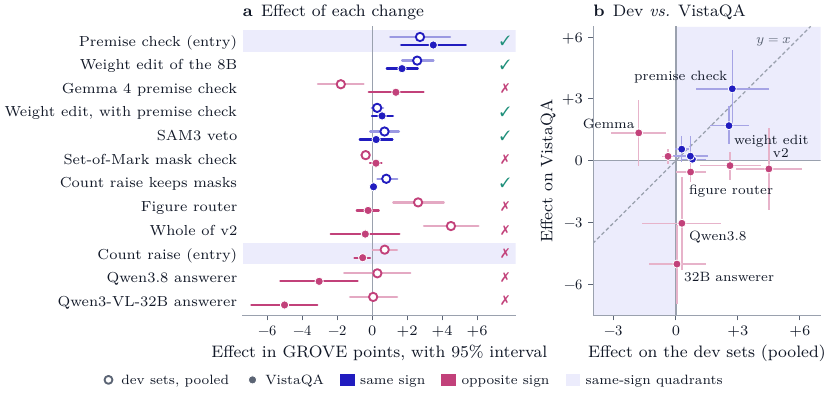}
\caption{Effect of each change on the dev sets and on VistaQA, in GROVE points with 95\% intervals
(Table~\ref{tab:transfer}). (a) The changes sorted by their VistaQA effect; hollow marks give the pooled dev-set effect and
filled marks the VistaQA effect, in indigo when the two signs agree (\agree) and in raspberry when they do not
(\disagree); shaded rows are the entry's two stages. (b) The same pairs as points; a change that transferred lies near the
diagonal, in a shaded quadrant where the two signs agree.}
\label{fig:transfer}
\end{figure}

\ishraq{This study answers RQ3 change by change: does an effect measured on the dev sets reappear on VistaQA?
Table~\ref{tab:transfer} puts each change's effect on the dev sets beside its effect on VistaQA, paired on the same
samples, and Figure~\ref{fig:transfer} draws the same pairs with their intervals.}

\ishraq{\textbf{The dev sets gave the right sign for 5 of 12 changes.} The correlation between pooled dev effects and VistaQA
effects is 0.23 (Pearson) and 0.08 (Spearman); the rows share systems and samples, so these numbers are descriptive only.
Only two changes have intervals that exclude zero with the same sign on both sides, the premise check and the weight edit,
and both lie close to the diagonal of Figure~\ref{fig:transfer}b. The largest dev gains, the whole of v2 and the figure
router, came to nothing, and the 32B answerer's level dev result hid a large loss. In the other direction, the Gemma
premise check and the Set-of-Mark check lost on the dev sets and did not lose on VistaQA. Proxy3 has no false-premise
questions, so the premise check can only lose there.}

\ishraq{\textbf{Changes decided on text-only probes.} Three more changes have no dev GROVE: thinking on reasoning questions
(MathVista 122 $\to$ 143 of 200; VistaQA v3 $+0.21$ over v2), the calculator (143 $\to$ 146; v4 $+0.16$ over v3) and 4\,MP
input (V*Bench 176 $\to$ 192 of 238; v5 $-0.34$ against v4). None of the three VistaQA differences is significant.}

\ishraq{\textbf{VistaQA is harder than every dev set.} Table~\ref{tab:across} shows the entry's chain on each set, and the
entry scores lower on VistaQA than on any dev set on every task and on both question types. On false-premise questions, the gap is about 21
points (70.9 against 91.5 and 92.2): the premise check catches 92\% and 96\% of them on the proxy development set and
proxy2 but 71\% on VistaQA, so the dev sets overstated its recall (Figure~\ref{fig:premise}b).}

\begin{table}[!htbp]
\centering\footnotesize
\setlength{\tabcolsep}{3pt}
\caption{Effect of each change in GROVE (AWQ judge). Columns 2--4 are point differences per dev set; the pooled and VistaQA
columns carry paired bootstrap 95\% CIs as subscripts, and the last column marks whether their signs agree (\agree) or not
(\disagree). Pooled uses the sets with a value in that row. Shaded, at the bottom: the entry's two stages.}
\label{tab:transfer}
\vspace{-2pt}%
\begin{tabular}{lrrrllc}
\toprule
\textbf{Change} & \ishraq{\textbf{Proxy}} & \textbf{Proxy2} & \textbf{Proxy3} & \textbf{Dev, pooled} & \textbf{VistaQA} & \textbf{Sign} \\
\midrule
Count raise keeps masks & -- & $+0.80$ & -- & $+0.80_{[+0.29,\,+1.43]}$ & $+0.07_{[+0.00,\,+0.17]}$ & \agree \\
Qwen3.8 for Qwen3.6, answer stage & $-1.19$ & $+1.72$ & -- & $+0.29_{[-1.61,\,+2.17]}$ & $-3.04_{[-5.26,\,-0.84]}$ & \disagree \\
Figure router, element evidence & -- & $+2.62$ & -- & $+2.62_{[+1.20,\,+4.09]}$ & $-0.24_{[-0.89,\,+0.37]}$ & \disagree \\
SAM3 veto ($\tau=0.4$) & $-0.32$ & $+1.67$ & -- & $+0.70_{[-0.12,\,+1.52]}$ & $+0.22_{[-0.72,\,+1.15]}$ & \agree \\
Whole of v2 \vs the entry & $+0.42$ & $+8.44$ & -- & $+4.50_{[+2.96,\,+6.07]}$ & $-0.40_{[-2.38,\,+1.55]}$ & \disagree \\
Weight edit of the 8B & $+3.24$ & $+1.92$ & -- & $+2.57_{[+1.71,\,+3.50]}$ & $+1.70_{[+0.83,\,+2.60]}$ & \agree \\
Weight edit, both with premise check & $+0.43$ & $+0.13$ & -- & $+0.28_{[-0.03,\,+0.62]}$ & $+0.56_{[-0.02,\,+1.18]}$ & \agree \\
Set-of-Mark mask check & $-0.48$ & $-0.38$ & $-0.10$ & $-0.38_{[-0.63,\,-0.15]}$ & $+0.21_{[-0.13,\,+0.53]}$ & \disagree \\
Qwen3-VL-32B answerer & $-2.01$ & $+1.42$ & $+1.87$ & $+0.05_{[-1.27,\,+1.42]}$ & $-5.02_{[-6.92,\,-3.15]}$ & \disagree \\
Gemma 4 31B premise check & $-2.19$ & $-0.73$ & $-3.82$ & $-1.80_{[-3.11,\,-0.50]}$ & $+1.35_{[-0.20,\,+2.93]}$ & \disagree \\
\midrule
\ours Premise check (8B) & $+7.20$ & $+1.83$ & $-7.17$ & $+2.73_{[+1.03,\,+4.45]}$ & $+3.49_{[+1.65,\,+5.35]}$ & \agree \\
\ours Count raise & $+1.54$ & $-0.10$ & -- & $+0.71_{[+0.02,\,+1.41]}$ & $-0.55_{[-1.02,\,-0.13]}$ & \disagree \\
\bottomrule
\end{tabular}
\vspace{-6pt}%
\end{table}

\begin{table}[!htbp]
\centering\footnotesize
\setlength{\tabcolsep}{5pt}
\caption{The entry's chain on each set: GROVE per task and question type, then text accuracy, mIoU and overall GROVE. Proxy3
has only spatial and comparison questions and no false premises; with no counting questions, its chain ends at the premise
check. VistaQA shaded.}
\label{tab:across}
\vspace{-2pt}%
\begin{tabular}{lrrrrrrrrrrr}
\toprule
 & \multicolumn{6}{c}{\textbf{Task}} & \multicolumn{2}{c}{\textbf{Question}} & \multicolumn{3}{c}{\textbf{Overall}} \\
\cmidrule(lr){2-7}\cmidrule(lr){8-9}\cmidrule(l){10-12}
\textbf{Set} & \textbf{Ident.} & \textbf{Attr.} & \textbf{OCR} & \textbf{Spat.} & \textbf{Count} & \textbf{Comp.} & \textbf{FP} &
\textbf{Gr.} & \textbf{Text} & \textbf{mIoU} & \textbf{GROVE} \\
\midrule
\ishraq{Proxy} (560) & 68.2 & 63.5 & 59.8 & 73.1 & 65.7 & 72.5 & 91.5 & 67.3 & 79.8 & 76.5 & 73.81 \\
Proxy2 (580) & 54.2 & 71.0 & 75.4 & 54.8 & 46.0 & 53.4 & 92.2 & 44.4 & 75.5 & 56.9 & 57.43 \\
Proxy3 (200) & -- & -- & -- & 73.3 & -- & 65.0 & -- & 70.8 & 81.5 & 67.7 & 70.78 \\
\midrule
\ours VistaQA (1,157) & 50.8 & 52.6 & 58.4 & 41.8 & 44.6 & 43.2 & 70.9 & 40.0 & 60.8 & 52.1 & 48.40 \\
\bottomrule
\end{tabular}
\vspace{-6pt}%
\end{table}

\end{document}